\documentclass[conference]{IEEEtran}
\usepackage{times}

\usepackage[numbers]{natbib}
\usepackage{multicol}
\usepackage{xcolor}

\usepackage{graphicx}
\usepackage{caption}
\usepackage{cuted}
\usepackage{amsmath}
\usepackage{algorithm} 
\usepackage{algpseudocode} 
\usepackage{booktabs} 
\usepackage{multirow} 
\usepackage{makecell} 
\definecolor{mylinkcolor}{RGB}{40, 115, 201}
\definecolor{mycitecolor}{RGB}{71, 191, 38}
\definecolor{TODO}{RGB}{255, 0, 0}
\definecolor{DGreen}{RGB}{107, 190, 35}

\newcommand{\argmin}{\operatornamewithlimits{argmin}}

\usepackage[bookmarks=true, colorlinks=true, citecolor=mycitecolor,linkcolor=mylinkcolor,urlcolor=mycitecolor]{hyperref}

\begin{document}


\title{RAMEN: Real-time Asynchronous Multi-agent Neural Implicit Mapping}

\author{Hongrui Zhao, Boris Ivanovic, and Negar Mehr}

\author{
\authorblockN{Hongrui Zhao}
\authorblockA{ICON Lab\\
Department of Aerospace Engineering\\
University of Illinois Urbana-Champaign\\
Email: hongrui5@illinois.edu}
\and
\authorblockN{Boris Ivanovic}
\authorblockA{NVIDIA Research\\
Email: bivanovic@nvidia.com}
\and
\authorblockN{Negar Mehr}
\authorblockA{ICON Lab\\
Department of Mechanical Engineering\\
University of California Berkeley\\
Email: negar@berkeley.edu}
}


%

\maketitle

\begin{strip}
  \begin{minipage}{\textwidth}\centering
    \vspace{-30pt}
        \includegraphics[trim={1.3cm 0cm 2.1cm 0cm},clip,width=0.85\textwidth]{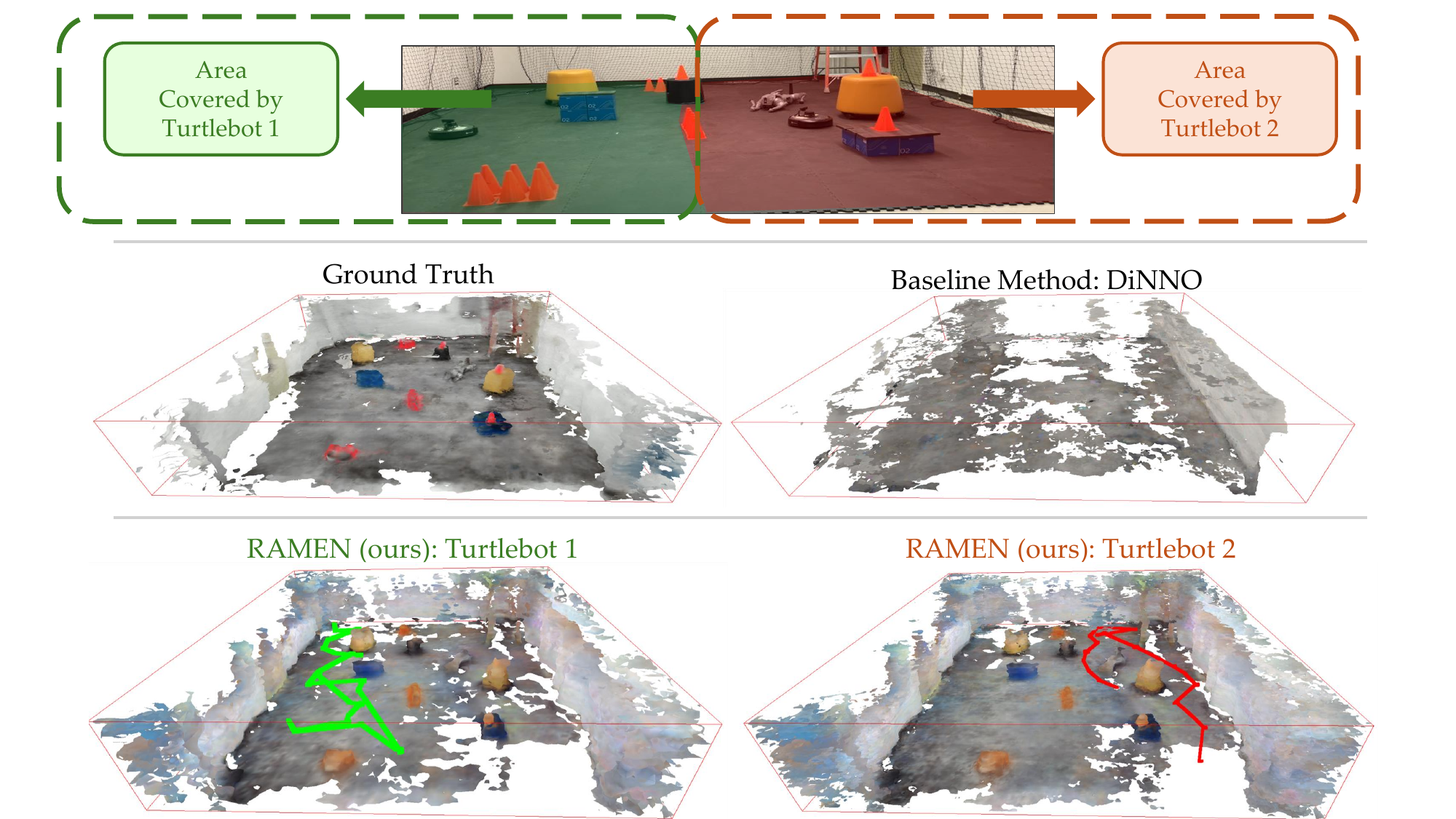}  
    \captionof{figure}{
    In a challenging real-world experiment with limited communication (agents can only exchange information every 30 seconds), our method RAMEN enables each turtlebot to successfully map the full scene while only physically visiting half of the scene (explored areas and trajectories are colored accordingly). Our method achieves accuracy comparable to the ground truth while the baseline method (DiNNO) fails to converge.
    }
    \label{fig:teaser}
  \end{minipage}
\end{strip}

\begin{abstract}
Multi-agent neural implicit mapping allows robots to collaboratively capture and reconstruct complex environments with high fidelity. 
However, existing approaches often rely on synchronous communication, which is impractical in real-world scenarios with limited bandwidth and potential communication interruptions. 
This paper introduces RAMEN: Real-time Asynchronous Multi-agEnt Neural implicit mapping, a novel approach designed to address this challenge.
RAMEN employs an uncertainty-weighted multi-agent consensus optimization algorithm that accounts for communication disruptions. 
When communication is lost between a pair of agents, each agent retains only an outdated copy of its neighbor's map, with the uncertainty of this copy increasing over time since the last communication.
Using gradient update information, we quantify the uncertainty associated with each parameter of the neural network map. 
Neural network maps from different agents are brought to consensus on the basis of their levels of uncertainty, with consensus biased towards network parameters with lower uncertainty.
To achieve this, we derive a weighted variant of the decentralized consensus alternating direction method of multipliers (C-ADMM) algorithm, facilitating robust collaboration among agents with varying communication and update frequencies.
Through extensive evaluations on real-world datasets and robot hardware experiments, we demonstrate RAMEN's superior mapping performance under challenging communication conditions. 
Supplementary videos and codes are available at \href{https://iconlab.negarmehr.com/RAMEN/}{https://iconlab.negarmehr.com/RAMEN/}.

\end{abstract}

\IEEEpeerreviewmaketitle

\section{Introduction}
Multi-robot systems rely on efficient communication and accurate map merging to explore and understand their environment. 
By sharing individual observations, agents gain a comprehensive understanding of the scene, enabling effective trajectory planning and task coordination. 
This capability is crucial for diverse applications like autonomous driving \cite{drive1,drive2,drive3}, warehouse automation \cite{warehouse}, and multi-UAV search and survey missions \cite{UAV}. Neural implicit maps, such as Neural Radiance Fields (NeRF) \cite{NeRF}, are emerging as a powerful tool for multi-robot mapping, offering a compelling alternative to traditional map representations \cite{CATNIPS1, CATNIPS2}. 
Neural implicit maps efficiently store complex geometries and realistic appearance information (including color and lighting) in compact neural networks, requiring minimal storage space on robots. 
Furthermore, high-level semantic information, such as object classes, can be readily integrated into neural implicit maps \cite{semanticNeRF, lerf2023}. 

However, challenges arise when multiple robots perform neural implicit mapping in real-time. 
While conventional map representations (e.g., point clouds, occupancy grids) are easily merged \cite{point_cloud, occupancy_map}, neural network parameters cannot be simply combined. 
Existing multi-agent neural implicit mapping methods often utilize distributed optimization algorithms, such as distributed stochastic gradient descent (DSGD) or C-ADMM, to gradually
bring different neural networks into \emph{consensus} \cite{DiNNO, MACIM, Di-NeRF}.
This consensus results in each robot possessing an identical neural network representing the collective sensor observations.
The consensus optimization, however, is vulnerable to divergence when communication is asynchronous.

Previous works \cite{ours} reveal that existing distributed optimization algorithms, such as the C-ADMM employed by the state-of-the-art multi-agent neural implicit mapping methods DiNNO~\cite{DiNNO} and Di-NeRF \cite{Di-NeRF}, are not robust to asynchronous communication or message loss.
These algorithms require perfect synchronization, where agents consistently receive the latest network parameters from neighbors before each optimization step. 
However, real-world robot communication is often asynchronous: updates may be delayed or lost due to bandwidth limitations, best-effort protocols (e.g., UDP), obstacles between agents, or agents moving out of range.
Consequently, as communication failures increase, multi-agent mapping can produce inaccurate reconstructions or even diverge entirely, as is the case for DiNNO~\cite{DiNNO} in Fig. \ref{fig:teaser}.
Existing approaches to handle such communication disruptions often rely on barriers that halt optimization to wait for the slowest agent \cite{asyn1, asyn2}. 
This is impractical for real-time robotic applications where continuous mapping is crucial for planning.

In this paper, we introduce RAMEN, Real-time Asynchronous Multi-agEnt Neural implicit mapping, to address these challenges.
RAMEN explicitly addresses the uncertainty inherent in distributed multi-agent systems by employing a frequency-based epistemic uncertainty quantification method. 
This method leverages gradient update information to assess the reliability of each agent's neural implicit map, recognizing that infrequent updates due to communication loss lead to higher uncertainty in the neural reconstruction. 
To achieve consensus between agents, we derive an uncertainty-weighted decentralized C-ADMM algorithm which enables RAMEN to effectively fuse knowledge from multiple robots while favoring more reliable information. 
To the best of our knowledge, this work presents the first real-world demonstration of multi-robot neural implicit mapping. 
Through extensive simulation and hardware experiments, we show that RAMEN successfully constructs maps in real-time despite communication disruptions, outperforming existing methods that fail under such conditions.
A summary of our key contributions is:
\begin{enumerate}
\def\labelenumi{\arabic{enumi}.}
\item We propose RAMEN, a multi-agent neural implicit mapping method with frequency-based uncertainty to address communication disruptions.
\item We derive an uncertainty-weighted decentralized C-ADMM algorithm, allowing RAMEN to prioritize reliable information during map consensus optimization.
\item We demonstrate the first real-world implementation of multi-robot neural implicit mapping, showcasing the robustness of RAMEN under challenging communication conditions.
\end{enumerate}

\section{Related Works} \label{section:related works}
\noindent \textbf{Real-time Neural Implicit Mapping}.
Neural implicit mapping methods based on NeRF utilize RGB or RGBD images to train neural networks for various scene prediction tasks, including color, volume density, occupancy, and signed distance.  
Examples include iSDF \cite{isdf} for real-time signed distance field reconstruction, H2-Mapping \cite{H2-Mapping} for efficient color and signed distance prediction on edge devices, and GO-Surf \cite{wang2022go-surf} for fast surface reconstruction using multi-resolution features.  
Some approaches, like iMAP \cite{iMAP} and NICE-SLAM \cite{NICE-SLAM}, integrate camera pose tracking with scene representation, while Co-SLAM \cite{wang2023coslam} further enhances reconstruction speed.  
However, these methods primarily focus on centralized training with a single agent.
In contrast, this paper builds upon Co-SLAM and introduces an asynchronous distributed training framework specifically designed for multi-robot systems.

\noindent\textbf{Multi-agent Mapping}. 
Existing multi-agent mapping methods involve transmitting either explicit local maps (e.g., 3D point clouds with image features \cite{indoor1,CP-SLAM, occupancy_map}) or raw sensor data to a central server for map alignment and fusion \cite{indoor2}.  
Alternatively, methods like MAANS \cite{MAANS}  employ a central server to fuse bird's-eye view images predicted by each agent.  
Recent approaches leverage neural implicit mapping to reduce communication costs by sharing compact neural networks instead of raw data \cite{DiNNO, Di-NeRF, MACIM}.
However, existing neural implicit mapping methods require synchronous communication, hindering their applicability in real-world robotic scenarios where communication is often asynchronous.  
Bandwidth limitations, best-effort protocols (e.g., UDP), and obstacles can lead to delayed or lost updates.
Our proposed RAMEN method addresses this gap by enabling robust multi-agent neural implicit mapping under asynchronous communication.

\noindent \textbf{Distributed Neural Network Optimization}.
Distributed optimization of neural networks allows multiple agents to collaboratively train a model without relying on a central server \cite{distributed}.
Common approaches include distributed stochastic gradient descent (DSGD) \cite{DSGD}, which combines local gradients with neighbors' parameters, and distributed stochastic gradient tracking (DSGT) \cite{DSGT}, which improves convergence by estimating the global gradient.
Other methods, like the decentralized linearized alternating direction method (DLM) \cite{DLM} and consensus alternating direction method of multipliers (C-ADMM) \cite{DiNNO},  optimize local objectives while enforcing agreement among agents. 
However, these methods typically require synchronous communication, limiting their practicality. 
While techniques like partial barriers and bounded delay \cite{asyn1, asyn2} allow for some degree of tolerance to communication disruptions by introducing waiting periods, they are primarily designed for distributed training with data servers and are not ideal for robotic tasks requiring continuous real-time updates.
RAMEN tackles communication challenges without relying on such barriers, enabling efficient and robust multi-robot collaborative learning.

\noindent \textbf{Uncertainty in Neural Implicit Maps.}
Quantifying uncertainty in neural implicit maps is crucial for tasks like next-best view selection, model refinement, and artifact removal.
Existing methods for estimating neural network parameter uncertainty often rely on computationally expensive techniques like deep ensembles \cite{ensemble} or variational Bayesian neural networks \cite{bayesian}, which require significant modifications to the training pipeline.
While alternative approaches estimate spatial uncertainty by perturbing input coordinates \cite{BayesRays} or network parameters \cite{ActiveNeuralMap}, they do not directly address the uncertainty of each learnable parameter.
In contrast, RAMEN efficiently quantifies parameter uncertainty by combining a frequency-based heuristic \cite{siming2024active} with a multi-resolution hash grid \cite{wang2023coslam}. 
This requires minimal extra computation and allows us to leverage uncertainty information for robust distributed mapping.

\section{Preliminaries}
This section provides the necessary background for our proposed approach. 
We first introduce Co-SLAM \cite{wang2023coslam}, a single-agent neural implicit mapping method that forms the foundation of RAMEN. 
We then discuss existing multi-agent neural implicit mapping approaches, assuming synchronous communications.
This discussion highlights the limitations of current methods and motivates the need for a system like RAMEN, which can handle the asynchronous communications common in real-world robotic scenarios.

\subsection{Single-agent Neural Implicit Mapping}
Co-SLAM, a single-agent neural implicit mapping method, represents scenes implicitly using three components: a multi-resolution hash-based feature grid \cite{instantNGP} which we denote by $V_\alpha = \{ V_\alpha^l \}^L_{l=1}$,
a geometry MLP decoder $F_\tau$, and a color MLP decoder $F_\phi$.
The feature grid $V_\alpha$ comprises $L$ levels, each storing feature vectors at grid vertices with resolutions increasing in a geometric progression from coarsest to finest.
Higher resolution levels have more vertices placed across the scene, thus capture finer details. 
We denote the learnable parameters of the feature grid and decoders as $\mathbf{\Theta}= \{ \alpha, \tau, \phi \}$.

Given the world coordinate $\mathbf{x}$ of a point, Co-SLAM's neural implicit map predicts its corresponding RGB value $\mathbf{c}$ and truncated signed distance function (SDF) value $s$.
The SDF value $s$ represents the minimum distance of a given point to the surface boundary of a geometric shape, with its sign indicating  whether the point lies inside (-) or outside (+) the surface.
To enhance the smoothness of the scene geometry,  one-blob encoding \cite{oneblob} $\gamma(\cdot)$ is applied to $\mathbf{x}$ before passing it to the geometry decoder $F_\tau$. 
With the encoded coordinate $\gamma(\mathbf{x})$ and feature vectors from the corresponding vertices $V_\alpha(\mathbf{x})$, the geometry decoder outputs a feature vector $\mathbf{h}$ and the SDF value $s$:
\begin{equation}
    F_\tau \left( \gamma(\mathbf{x}),  V_\alpha(\mathbf{x}) \right) \to (\mathbf{h}, s).
    \label{eq:geometry}
\end{equation}
The color decoder then outputs the RGB value $\mathbf{c}$:
\begin{equation}
    F_\phi \left( \gamma(\mathbf{x}),  \mathbf{h}) \right) \to \mathbf{c}.
    \label{eq:color}
\end{equation}

Co-SLAM renders RGB and depth images from its neural implicit map. 
With the camera intrinsic calibration matrix $K$ and pose $P$, each pixel with 2D image plane coordinates $(u,v)$ is associated with a ray $\mathbf{r}$ in the direction $PK^{-1}(u,v,1)^\top$.
Given the camera origin $\mathbf{o}$ and ray direction $\mathbf{r}$, we sample $M$ points $\mathbf{x}_i = \mathbf{o} + d_i \mathbf{r}$ along the ray, where $d_i$ is the distance from the camera origin.  
For each point $\mathbf{x}_i$, we obtain the corresponding $s_i$ and $\mathbf{c}_i$ using \eqref{eq:geometry} and \eqref{eq:color}.
The pixel color $\hat{\mathbf{c}}$ and depth $ \hat{d}$ are then rendered as:
\begin{subequations}
    \begin{align}
        \hat{\mathbf{c}} = \frac{1}{\sum^M_{i=1} w_i } \sum_{i=1}^M w_i \mathbf{c}_i,  \\
        \hat{d} = \frac{1}{\sum^M_{i=1} w_i } \sum_{i=1}^M w_i \mathbf{d}_i,
    \end{align}
\end{subequations}
where $w_i$ represents the weight assigned to each sampled point along the ray, peaking at the surface of a geometric shape \cite{neuralRGBD}. 
Given a truncated distance $tr$ (we truncate the SDF value $s$ to $tr$ for points far from the surface), we compute $w_i$ with two Sigmoid functions $\sigma(\cdot)$:
\begin{equation}
    w_i = \sigma\left(\frac{s_i}{tr}\right) \sigma\left(-\frac{s_i}{tr}\right).
\end{equation}

We train the neural implicit map by minimizing the difference between rendered RGBD images and the images from robot's camera.
We use notation $R$ to denote the actual images collected by the robot.
The objective function is then given as:
\begin{equation}
    L^{obj}(\mathbf{\Theta}, R) = L_{rgb} + L_d + L_{sdf} + L_{fs} + L_{smooth},
    \label{eq:obj}
\end{equation}
where $L_{rgb}$ and $L_d$ are L2 losses between the rendered and observed RGB and depth images, respectively; $L_{sdf}$ is the loss for SDF values; $L_{fs}$ encourages points far from the surface to have a SDF value equal to the truncated distance $tr$, and $L_{smooth}$ promotes smoothness in the reconstructed geometry.
For a detailed explanation of each loss term, please refer to \cite{wang2023coslam}.

\subsection{Synchronous Multi-agent Mapping}
We now discuss how Co-SLAM can be extended into synchronous multi-agent mapping.
In a multi-agent setup, a group of agents collaboratively map their environment and share information within a communication graph $G=(\mathcal{V}, \mathcal{E})$. 
Each agent is represented as a node $i \in \mathcal{V}$, and their communication connectivity is defined by a set of edges $\mathcal{E}$.
Each agent $i$ can communicate with its neighbors $N_i = \{ j  \in \mathcal{V} \mid (i,j) \in \mathcal{E} \}$, sharing its learned parameters $\mathbf{\Theta}_i$ instead of raw images.  
Each agent $i$ captures posed RGBD images $R_i$ with its onboard camera. 
During each mapping iteration, agent $i$ samples pixels from $R_i$ and minimizes its local objective function $L^{obj}_i$:
\begin{equation}
    L^{obj}_i(\mathbf{\Theta}_i, R_i) = {L_{rgb}}_i + {L_d}_i + {L_{sdf}}_i + {L_{fs}}_i + {L_{smooth}}_i,
    \label{eq:obj_i}
\end{equation}
To achieve consensus, each agent also aligns its model parameters $\mathbf{\Theta}_i$ with those of its neighbors $\mathbf{\Theta}_j$, for all $j \in N_i$.
Through consensus, agents converge on a shared neural map, allowing them to exchange environmental information.
This problem is formulated as minimizing the sum of local objectives while ensuring agreement among the neural maps~\cite{decentralizedCADMM}:
\begin{subequations}
    \label{eq:decentralized} 
    \begin{align}
        &\min_{ \{\mathbf{\Theta}_1, \cdots\}  \{\mathbf{z}_{ij},\cdots\} } \; \sum_{i \in \mathcal{V}}  L^{obj}_i(\mathbf{\Theta}_i, R_i),   \\
        &\text{s.t.} \; \mathbf{\Theta}_i=\mathbf{z}_{ij}, \mathbf{\Theta}_j=\mathbf{z}_{ij}, \; \forall (i,j) \in \mathcal{E}, 
    \end{align}
\end{subequations}
where $\mathbf{z}_{ij}$ is a local auxiliary variable imposing consensus on neighboring agents $i$ and $j$. 
This formulation represents a \emph{decentralized consensus problem} (or graph consensus problem), where agreement is enforced only between pairs of neighbors. 
In contrast, a \emph{global consensus problem} requires all $\mathbf{\Theta}_i$ to agree with a global variable $\mathbf{z}$, managed by a central fusion center \cite{globalCADMM}. 
Decentralized consensus is more suitable for multi-robot systems, as it avoids the single point of failure and vulnerability associated with a fusion center.

To solve \eqref{eq:decentralized}, we first formulate the augmented Lagrangian:
\begin{align}
&\mathcal{L}^{a} = \sum_{i \in \mathcal{V}} \Bigg( L^{obj}_i(\mathbf{\Theta}_i, R_i) + \sum_{j \in N_i} \boldsymbol{\lambda} _{ij}^\top(\mathbf{\Theta}_i - \mathbf{z}_{ij}) \nonumber \\ 
&+\mathbf{v}_{ij}^\top(\mathbf{\Theta}_j - \mathbf{z}_{ij}) + \frac{\rho}{2} \Vert \mathbf{\Theta}_i - \mathbf{z}_{ij}\Vert_2^2 + \frac{\rho}{2}\Vert \mathbf{\Theta}_j - \mathbf{z}_{ij}\Vert_2^2  \Bigg), \label{eq:augLa}
\end{align}
where $\boldsymbol{\lambda} _{ij}$ and $\mathbf{v}_{ij}$ are dual variables (or Lagrange multipliers) penalizing violations of consensus constraints and bringing $\mathbf{\Theta}_i$ and $\mathbf{\Theta}_j$ to $\mathbf{z}_{ij}$,
$\rho$ is a positive penalty parameter which determines how fast we want to achieve consensus, and
$\Vert \cdot \Vert_2$ is L2 norm.
Additionally, we define $\mathbf{p}_i = \sum_{j \in N_i} \boldsymbol{\lambda} _{ij} + \mathbf{v}_{ji}$, which combines all dual variables associated with $\mathbf{\Theta}_i$.

Multi-agent mapping employs an iterative algorithm to continually update neural implicit maps. 
Superscript $t$ is used to denote the values of variables at iteration $t$.
For agent $i$, at mapping iteration $t$, we apply ADMM \cite{globalCADMM} to update $\mathbf{\Theta}_i$ and $\mathbf{p}_i$ in an alternating fashion to minimize \eqref{eq:augLa}:
\begin{subequations}
    \begin{align}
        \mathbf{\Theta}_i^{t+1} =& \argmin_{\mathbf{\Theta}_i} \; L^{obj}_i(\mathbf{\Theta}_i, R_i) + \mathbf{\Theta}_i^\top\mathbf{p}_i^t \nonumber \\
        & + \rho \sum_{j \in N_i} \left \Vert  \mathbf{\Theta}_i - \frac{\mathbf{\Theta}_i^t + \mathbf{\Theta}_j^t}{2} \right \Vert_2^2,  \label{eq:primal_update} \\
        \mathbf{p}_i^{t+1} =& \mathbf{p}_i^t + \rho \sum_{j \in N_i} ( \mathbf{\Theta}_i^{t+1} - \mathbf{\Theta}_j^{t+1}).  \label{eq:dual_update}
    \end{align}
\label{eq:ADMM}
\end{subequations}

We refer to \eqref{eq:primal_update} as the \emph{primal update} and \eqref{eq:dual_update} as the \emph{dual update}. 
Instead of performing the exact minimization in the primal update, we approximate $\mathbf{\Theta}_i^{t+1}$ by taking few steps of stochastic gradient descent.
This approach, used in DiNNO \cite{DiNNO} and Di-NeRF \cite{Di-NeRF} for synchronous multi-agent mapping, requires continuous communication between agents $i$ and $j$ if $(i,j) \in \mathcal{E}$ initially. 
This ensures that the latest $\mathbf{\Theta}_i$ and $\mathbf{\Theta}_j$ are always available for \eqref{eq:ADMM}. 
However, this requirement is impractical for real-world robotic systems. 
In the next section, we introduce an uncertainty-weighted version of \eqref{eq:ADMM} to address this limitation.

\section{Asynchronous Multi-agent Neural Implicit Mapping}
This section details the core components of RAMEN's asynchronous multi-agent mapping framework. 
In our context, asynchronism arises from two primary sources: communication drops and variations in processing speed. 
This work primarily addresses communication drops, as they pose a significant challenge for both homogeneous and heterogeneous robot teams.
We first derive an uncertainty-weighted version C-ADMM algorithm, utilizing uncertainty to favor reliable information in the consensus optimization process.
Then, we introduce our proposed frequency-based epistemic uncertainty quantification method, detailing how it assesses the reliability of each agent's contributions to the collaborative mapping process.

\subsection{Uncertainty-Weighted Decentralized C-ADMM} \label{subsection:weighted C-ADMM}
Asynchronous communication presents two key challenges:
\begin{enumerate}
    \item {\emph{Mitigating the influence of uncertain parts of outdated models}:}
    When communication between agents $i$ and $j$ is disrupted, agent $i$ at iteration $t$ may possess an outdated copy of agent $j$'s model parameters, denoted as $\mathbf{\Theta}_j^{t_{old}}$, where $t_{old} \ll t$. 
    This outdated model may represent some regions of the environment with high uncertainty and poor quality due to lack of mapping updates.
    Therefore, it is crucial to limit the influence of uncertain parts of $\mathbf{\Theta}_j^{t_{old}}$ on agent $i$'s current model $\mathbf{\Theta}_i^{t}$ to avoid degrading its accuracy.

    \item{\emph{Leveraging valuable information from outdated models}:}
    Despite being outdated, $\mathbf{\Theta}_j^{t_{old}}$ may still contain valuable information for agent $i$, such as details about regions explored by agent $j$ but not yet visited by agent $i$.  
    We want to identify the neural network parameters that encode these information and assign them higher weights when enforcing consensus.
\end{enumerate}

To address these challenges, we derive an uncertainty-weighted decentralized C-ADMM algorithm that prioritizes reliable neural network parameters during the consensus optimization process.
While various weighted C-ADMM algorithms exist \cite{uncertaintyCAMM, Ling2016WeightedAF, edge_CADMM_1, edge_CADMM_2}, this paper introduces, to the best of our knowledge, the first decentralized C-ADMM algorithm that individually weights each optimization parameter according to its associated uncertainty.
This allows for fine-grained control over the influence of each parameter during the consensus process, enabling more robust and accurate multi-agent mapping.

Following \cite{uncertaintyCAMM}, we start by writing the weighted augmented Lagrangian based on \eqref{eq:augLa}:
\begin{align}
&\mathcal{L}^{wa} = \sum_{i \in \mathcal{V}} \Bigg( L^{obj}_i(\mathbf{\Theta}_i, R_i) + \sum_{j \in N_i} \boldsymbol{\lambda} _{ij}^\top(\mathbf{\Theta}_i - \mathbf{z}_{ij}) \nonumber  \\
&  +\mathbf{v}_{ij}^\top(\mathbf{\Theta}_j - \mathbf{z}_{ij}) + \frac{\rho}{2} \Vert \mathbf{\Theta}_i - \mathbf{z}_{ij}\Vert_{\mathbf{W}_{ij}}^2 + \frac{\rho}{2}\Vert \mathbf{\Theta}_j - \mathbf{z}_{ij}\Vert_{\mathbf{W}_{ji}}^2 \Bigg),  \label{eq:waLa}
\end{align}
where we use a weighted norm $\Vert \mathbf{x} \Vert^2_{\mathbf{W}} = \mathbf{x}^\top \mathbf{W} \mathbf{x} $ instead of L2 norm in \eqref{eq:augLa}. 
Note that $\mathbf{W}_{ij}^t, \mathbf{W}_{ji}^t$ are weight matrices reflecting the relative reliability of information shared between agents i and j at iteration t.  
Intuitively, a higher weight indicates greater confidence/lower uncertainty in the corresponding information.

Applying ADMM \cite{globalCADMM} to \eqref{eq:waLa}, at each iteration $t$ for agent $i$, we alternatively update local consensus variable $\mathbf{z}_{ij}$, neural implicit map $\mathbf{\Theta}_i$, and consensus-enforcing dual variables $\boldsymbol{\lambda} _{ij}, \mathbf{v}_{ij}$.
We obtain $\mathbf{z}_{ij}^{t+1}$ by  setting the derivative of \eqref{eq:waLa} with respect to $\mathbf{z}_{ij}$ to zero:
\begin{align}
    &\mathbf{z}_{ij}^{t+1} = (\mathbf{W}_{ij}^t+\mathbf{W}_{ji}^t)^{-1}(\mathbf{W}_{ij}^t\mathbf{\Theta}_i^t + \mathbf{W}_{ji}^t\mathbf{\Theta}_j^t) \nonumber \\
    & \hspace{1.5cm} + \frac{1}{\rho}(\mathbf{W}_{ij}^t+\mathbf{W}_{ji}^t)^{-1}(\boldsymbol{\lambda} _{ij}^t + \mathbf{v}_{ij}^t ). \label{eq:z_ij}
\end{align}


After obtaining $\mathbf{z}_{ij}^{t+1}$ and $\mathbf{\Theta}_i^{t+1}$, we compute the dual variables $\boldsymbol{\lambda} _{ij}^{t+1}, \mathbf{v}_{ij}^{t+1}$ by performing gradient ascents to penalize consensus violations:
\begin{subequations}
    \begin{align}
        &\boldsymbol{\lambda} _{ij}^{t+1} = \boldsymbol{\lambda} _{ij}^{t} + \rho \mathbf{W}_{ij}^t \nabla_{\boldsymbol{\lambda} _{ij}}\mathcal{L}^{wa},  \label{eq:grad_ascent_lambda} \\
        &\mathbf{v}_{ij}^{t+1} = \mathbf{v}_{ij}^{t} + \rho \mathbf{W}_{ji}^t \nabla_{\mathbf{v}_{ij}}\mathcal{L}^{wa} \label{eq:grad_ascent_v},
    \end{align}
    \label{eq:grad_ascent}
\end{subequations}
where the gradients are:
\begin{subequations}
    \begin{align}
        &\nabla_{\boldsymbol{\lambda} _{ij}}\mathcal{L}^{wa} = \mathbf{\Theta}_i^{t+1} - \mathbf{z}_{ij}^{t+1}, \\
        &\nabla_{\mathbf{v}_{ij}}\mathcal{L}^{wa} = \mathbf{\Theta}_j^{t+1} - \mathbf{z}_{ij}^{t+1},
    \end{align}
\end{subequations}
In \eqref{eq:grad_ascent}, $\mathbf{W}_{ij}^t$ and $\mathbf{W}_{ji}^t$ apply strong penalty if $\mathbf{\Theta}_i^{t+1}$ and $\mathbf{\Theta}_j^{t+1}$ diverge from the reliable information in $\mathbf{z}_{ij}^{t+1}$.

Initializing the dual variables as $\boldsymbol{\lambda} _{ij}^0 = \mathbf{v}_{ij}^0 = \mathbf{0}$, we can observe that $\boldsymbol{\lambda} _{ij}^t = -\mathbf{v}_{ij}^t$ by combining \eqref{eq:z_ij}, \eqref{eq:grad_ascent_lambda}, and \eqref{eq:grad_ascent_v}.
We also notice  $\boldsymbol{\lambda} _{ij}^t = \mathbf{v}_{ji}^t$.
This allows us to rewrite \eqref{eq:z_ij} as:
\begin{equation}
    \mathbf{z}_{ij}^{t+1} = (\mathbf{W}_{ij}^t+\mathbf{W}_{ji}^t)^{-1}(\mathbf{W}_{ij}^t\mathbf{\Theta}_i^t + \mathbf{W}_{ji}^t\mathbf{\Theta}_j^t)
    \label{eq:final_z},
\end{equation}
Note that~\eqref{eq:final_z} shows $\mathbf{z}_{ij}^{t+1}=\mathbf{z}_{ji}^{t+1}$, and \eqref{eq:final_z} can be interpreted as a weighted average between $\mathbf{\Theta}_i^t$ and $\mathbf{\Theta}_j^t$, where higher weights are assigned to parameters with lower uncertainty. Thus, $\mathbf{z}_{ij}^{t+1}$ weights the parameters based on how reliable they are, serving as the consensus target for both agents.

\begin{figure}[hbt!]
	\centerline{\includegraphics[trim={0cm 0cm 0cm 0cm},clip,width=\columnwidth]{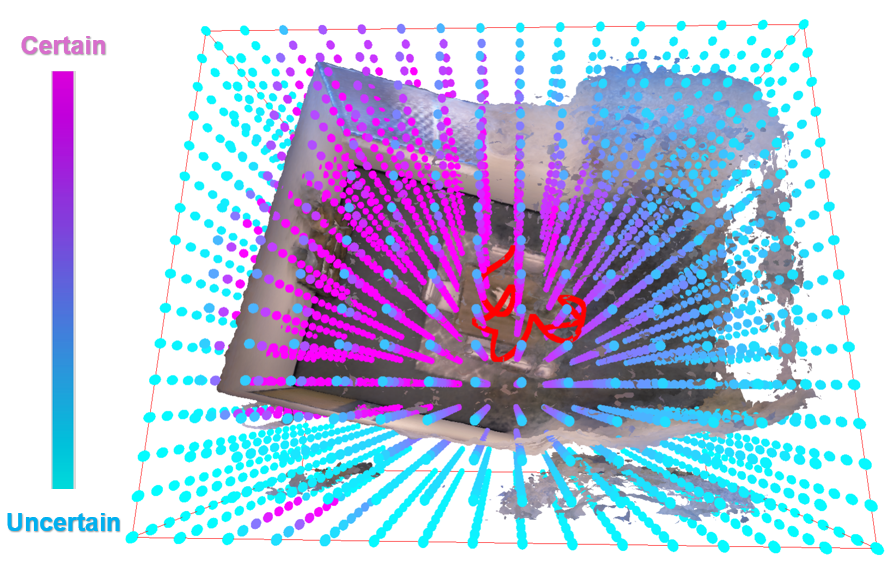}} 
	\caption{This visualization shows the uncertainty associated with the parameters in the coarsest feature grid $V_\alpha^{l=1}$ of the neural implicit map.
             We represent this uncertainty by overlaying color-coded vertices on the reconstructed 3D mesh. 
             The robot's trajectory is shown in red. The robot's trajectory and field-of-view are limited to the left half of the room.
             As expected, high uncertainty (blue vertices) corresponds to the right half of the map not explored by the robot where the reconstructed geometry is inaccurate.
    }
	\label{fig:uncertainty}
\end{figure}

To compute gradients for obtaining $\mathbf{\Theta}_i^{t+1}$, we are interested in all terms associated with $\mathbf{\Theta}_i$ (including $\mathbf{v}_{ji}^\top \mathbf{\Theta}_i = \boldsymbol{\lambda} _{ij}^\top \mathbf{\Theta}_i $ and $\frac{\rho}{2}\Vert \mathbf{\Theta}_i - \mathbf{z}_{ji}\Vert_{\mathbf{W}_{ij}}^2 = \frac{\rho}{2}\Vert \mathbf{\Theta}_i - \mathbf{z}_{ij}\Vert_{\mathbf{W}_{ij}}^2$ from the $j^{\text{th}}$ element of $\mathcal{L}^{wa}$).
We find $\mathbf{\Theta}_i^{t+1}$ that minimizes \eqref{eq:waLa} to be:
\begin{align}
    &\mathbf{\Theta}_{i}^{t+1} = \argmin_{\mathbf{\Theta}_i} \; L^{obj}_i(\mathbf{\Theta}_i, R_i) + \sum_{j \in N_i}  2\mathbf{\Theta}_i^\top \boldsymbol{\lambda} _{ij}^t \nonumber \\
    & +\rho \Vert \mathbf{\Theta}_i - (\mathbf{W}_{ij}^t+\mathbf{W}_{ji}^t)^{-1}(\mathbf{W}_{ij}^t\mathbf{\Theta}_i^t + \mathbf{W}_{ji}^t\mathbf{\Theta}_j^t) \Vert_{\mathbf{W}_{ij}^t}^2
    \label{eq:weight_primal},
\end{align}
This gives us the primal update of the weighted decentralized C-ADMM.
By defining $\mathbf{p}_i^t := 2 \sum_{j \in N_i} \boldsymbol{\lambda} _{ij}^t$, we can rewrite \eqref{eq:weight_primal} as:
\begin{align}
    &\mathbf{\Theta}_{i}^{t+1} = \argmin_{\mathbf{\Theta}_i} \; L^{obj}_i(\mathbf{\Theta}_i, R_i) + \mathbf{\Theta}_i^\top \mathbf{p}_i^t \nonumber \\
    & + \rho\sum_{j \in N_i}  \Vert \mathbf{\Theta}_i - (\mathbf{W}_{ij}^t+\mathbf{W}_{ji}^t)^{-1}(\mathbf{W}_{ij}^t\mathbf{\Theta}_i^t + \mathbf{W}_{ji}^t\mathbf{\Theta}_j^t) \Vert_{\mathbf{W}_{ij}^t}^2
    \label{eq:weight_primal_rewrite},
\end{align}
Using \eqref{eq:grad_ascent_lambda}, we perform dual update to get $\mathbf{p}_i^{t+1}$:
\begin{align}
    &\mathbf{p}_i^{t+1} = \mathbf{p}_i^t + \nonumber \\
    &2\rho\mathbf{W}_{ij}^t \sum_{j \in N_i}   (\mathbf{W}_{ij}^t+\mathbf{W}_{ji}^t)^{-1}(\mathbf{W}_{ji}^t\mathbf{\Theta}_i^{t+1} - \mathbf{W}_{ji}^t\mathbf{\Theta}_j^{t+1}),
    \label{eq:weigt_dual}
\end{align}
After going into the next iteration, we need to update the weights which reflect how reliable the network parameters are. We will discuss how we can update the weights in the next subsection. 


\begin{figure}[hbt!]
	\centerline{\includegraphics[trim={0cm 2.5cm 3.5cm 1.5cm},clip,width=\columnwidth]{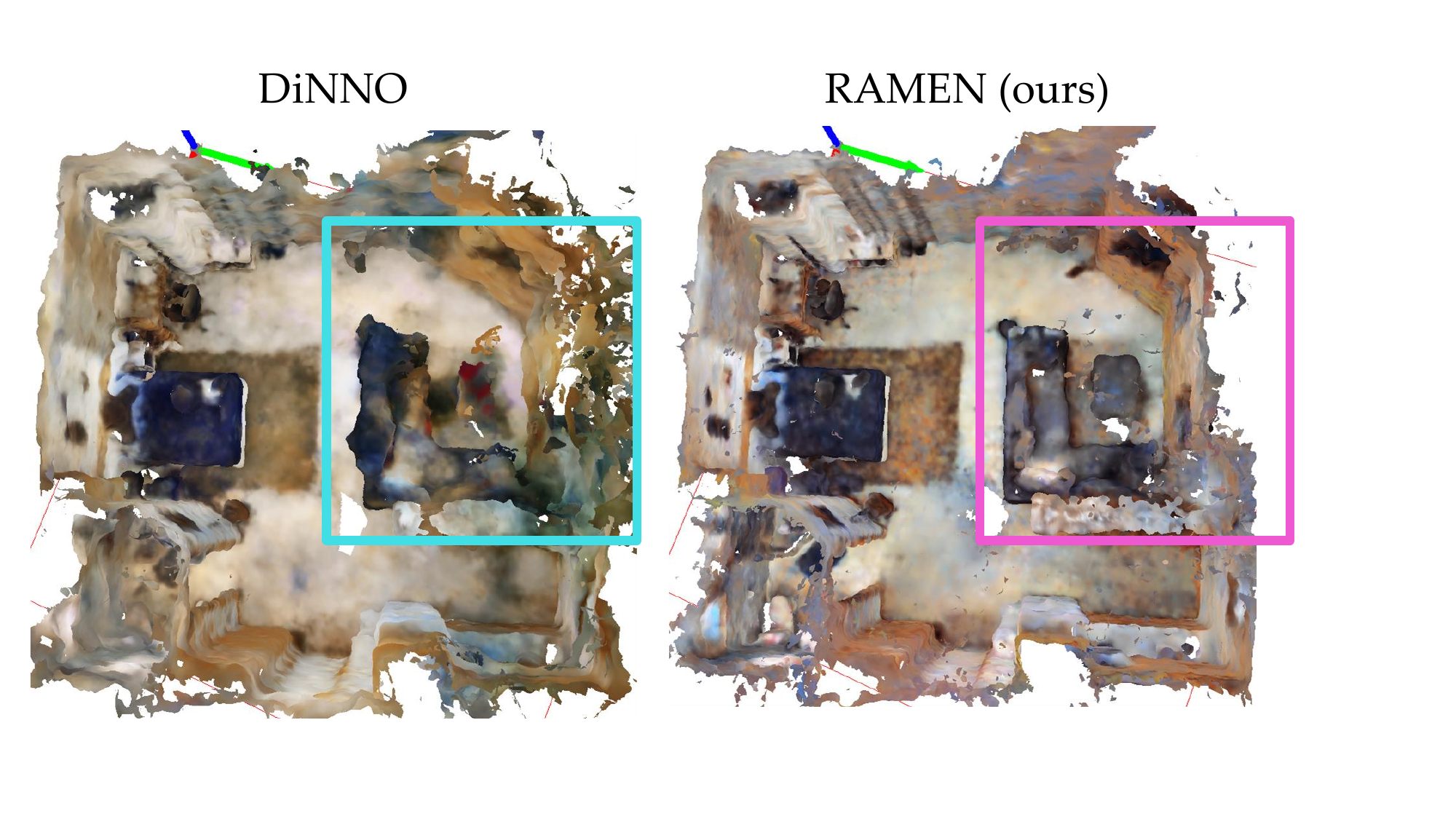}} 
	\caption{Comparison of scene reconstructions from DiNNO and RAMEN (ours) on ScanNet ``scene0000" with only 50\% communication success rate. RAMEN produces more accurate and detailed geometry in the highlighted region.}
	\label{fig:sim_test1}
\end{figure}

\subsection{Frequency-based Epistemic Uncertainty}

We quantify the \emph{epistemic} uncertainty associated with each model parameter to obtain weights for our C-ADMM algorithm. 
Epistemic uncertainty reflects the model's lack of knowledge about the environment it is mapping \cite{BayesRays}.
We base our simple-yet-effective uncertainty quantification method on two key observations:
\begin{enumerate}
    \item 
    The uncertainty of an area decreases monotonically with the frequency of observations \cite{siming2024active}.
    \item 
    Unlike the highly correlated parameters in a standard MLP, each feature vector in the feature grid $V_\alpha$ corresponds to a unique vertex in the grid \cite{BayesRays}. 
    Updates to a vertex primarily occur when the robot visits its corresponding spatial location \cite{instantNGP}.
\end{enumerate}
Based on these two observations, \emph{we associate the number of gradient updates a feature grid parameter receives with its epistemic uncertainty}.  
More updates indicate more frequent visits to the corresponding location, implying lower uncertainty. 
Due to the correlated nature of parameters in MLP decoders $F_\tau, F_\phi$ and their lack of direct spatial correspondence, we pretrain the decoders using the NICE-SLAM apartment dataset \cite{NICE-SLAM} and keep their parameters fixed during mapping.
Thus, unlike Co-SLAM, we only optimize the parameters $\alpha$ in the feature grid $V_\alpha$, resulting in $\mathbf{\Theta}= \{ \alpha \}$ with $n$ learnable parameters.
We flatten $\mathbf{\Theta}$ into a $n \times 1$ vector in our computations.

Our proposed frequency-based epistemic uncertainty quantifier is formulated as:
\begin{equation}
     \mathbf{u}_i^t = \sum_{k=0}^t \text{sgn} \left( \text{abs} \left( \frac{\partial  L^{obj}_{i,k} }{\partial \mathbf{\Theta}_i^{k}}  \right) \right),
     \label{eq:uncertainty}
\end{equation}
where $\mathbf{u}_i^t$, a $n \times 1$ vector, represents the uncertainty associated with each feature grid parameter at iteration $t$.
$L^{obj}_{i,k}$ is the Co-SLAM reconstruction objective loss \eqref{eq:obj_i} of agent $i$ at iteartion $k$;  
$\text{abs}(\cdot)$ is the absolute value function; 
$\text{sgn}(\cdot)$ is the sign function returning 1 when the given value $>0$ and returning 0 otherwise. 
$\mathbf{u}_i^t$ tracks the count of non-zero gradient updates for each parameter, establishing a link between the frequency of spatial location visits and their uncertainty.
Fig \ref{fig:uncertainty} visualizes the uncertainty of the parameters in the coarsest feature grid $V_\alpha^{l=1}$ in an example using our proposed method, demonstrating that our method accurately captures the uncertainty of the neural implicit map.

\begin{algorithm} 
	\caption{RAMEN: Real-time Asynchronous Multi-agent Neural Implicit Mapping}
	\label{procedure}
	\begin{algorithmic}[1]
		
		\For{$i \in \mathcal{V}$}
			\State $\mathbf{p}_i^{0} = 0$
			\State $\mathbf{\Theta}_i^0 = \mathbf{\Theta}_{\text{initial}} $ 
		\EndFor 

		\While {true}
			\For{$i \in \mathcal{V}$}
                \State Capture images and add to database $R_i$
				\State Send $\mathbf{\Theta}_i^t$ and $\mathbf{u}_i^t$ to neighbors $N_i$
			\EndFor 

			\For{$i \in \mathcal{V}$}
                \State Compute \eqref{eq:weight} to get $\mathbf{W}_{ij}^{t}, \mathbf{W}_{ji}^{t}$
                \For{$b \leftarrow 0 \; \text{to} \; B$}
                    \State Sample from $R_i$
                    \State Gradient descent on \eqref{eq:weight_primal_rewrite} to approximate  $\mathbf{\Theta}_{i}^{t+1}$
                \EndFor
				\State Perform dual variable update \eqref{eq:weigt_dual} to obtain $\mathbf{p}_i^{t+1}$
                \State Compute \eqref{eq:uncertainty} to get $\mathbf{u}_i^{t+1}$
			\EndFor 

        \EndWhile
	\end{algorithmic}
\end{algorithm}

Since the number of gradient updates could be infinite, we cannot directly use $\mathbf{u}_i^t$ as weights for consensus optimization in \ref{subsection:weighted C-ADMM}. 
Simply normalizing $\mathbf{u}_i^{t_i}$ and $\mathbf{u}_j^{t_j}$ independently to a range of $[0, 1]$ would eliminate the proportional relationship between their elements.
For example, the maximum value in $\mathbf{u}_i^{t_i}$ might be substantially larger than the maximum in $\mathbf{u}_j^{t_j}$, a difference that would be lost after independent normalization.
To preserve these proportional relationships and ensure that the weights reflect the relative uncertainties, we employ a linear scaling and shifting operation.
Given $\mathbf{u}_i^{t_i}$ and $\mathbf{u}_j^{t_j}$ for agent $i$ and its neighbor $j$, we obtain $n \times n$ weight matrices $\mathbf{W}_{ij}$ and $\mathbf{W}_{ji}$ as follows:
\begin{subequations}
   \label{eq:weight}
   \begin{gather}
        \mathbf{u}_{sum}:= \mathbf{u}_i^{t_i} + \mathbf{u}_j^{t_j}, \\
        \epsilon = \frac{\beta_{u} - \beta_{l}}{\text{max}(\mathbf{u}_{sum}) - \text{min}(\mathbf{u}_{sum}) } \label{eq:episilon}, \\
        \zeta = \beta_{l} - \epsilon \cdot \text{min}(\mathbf{u}_{sum}) \label{eq:zeta}, \\
        \mathbf{W}_{ij} = \text{diag}(\epsilon \cdot \mathbf{u}_i^{t_i} + \zeta), \mathbf{W}_{ji} = \text{diag}(\epsilon \cdot \mathbf{u}_j^{t_j} + \zeta),
    \end{gather}
\end{subequations}
where $\beta_{u}$ and $\beta_{l}$ are the user-specified upper and lower bounds of the weights;
$\text{max}(\cdot)$ and $\text{min}(\cdot)$ return the maximum and the minimum elements of the input vector;
$\text{diag}(\cdot)$ returns a $n \times n$ matrix using the input as its diagonal elements;
$\epsilon$ \eqref{eq:episilon} and $\zeta$ \eqref{eq:zeta} are scaling and shifting terms for normalization.
The normalization \eqref{eq:weight} preserves the proportional relationship between the uncertainty values of agents $i$ and $j$ by applying the same $\epsilon$
and $\zeta$ to both $\mathbf{u}_i^{t_i}$ and $\mathbf{u}_j^{t_j}$.
These weight matrices are then incorporated into the weighted decentralized C-ADMM algorithm to effectively leverage information from neighboring models while accounting for their uncertainty.

The entire multi-agent mapping process of RAMEN is summarized in Algorithm \ref{procedure}.
All agents use the same initialization $\mathbf{\Theta}_{\text{initial}}$ for their neural implicit maps.
Note that, instead of the exact minimization in \eqref{eq:weight_primal_rewrite}, we obtain $\mathbf{\Theta}_i^{t+1}$ by taking $B$ steps of stochastic gradient descent, where $B$ is a user-defined constant. 

\section{Results}
We evaluate RAMEN's performance through simulation and hardware experiments. 
For simulation, we use two datasets: Replica \cite{replica19arxiv}, containing synthetic posed RGBD images of various indoor scenes, and ScanNet \cite{dai2017scannet}, which offers challenging real-world posed RGBD images of large indoor spaces. 
We benchmark RAMEN against three state-of-the-art baselines: DSGD \cite{DSGD}, DSGT \cite{DSGT}, and DiNNO \cite{DiNNO}, detailed in Section \ref{section:related works}.
To simulate communication disruptions, we allow only a specified percentage of messages to be successfully transferred between agents.

\begin{table*}[!hbt]
  \caption{RAMEN significantly outperforms prior state-of-the-art approaches for multi-agent neural implicit mapping in challenging communication conditions.}
  \centering
  \resizebox{2\columnwidth}{!}{%
    \begin{tabular}{ccrrrrr}
      \toprule
       Method & Metric & office1 & room1 & scene0000 &  scene0106 & Avg.  \\
      \midrule
      \multirow{3}{*}{DSGD} & \emph{Artifacts} (cm) $\downarrow$ & 6.38$\pm$1.76 & 16.39$\pm$0.45 & 14.59$\pm$0.57 & 100.40$\pm$5.02 & 34.34$\pm$1.95\\
                            & \emph{Holes} (cm) $\downarrow$ & 2.70$\pm$0.20 & 2.55$\pm$0.02 & 5.03$\pm$0.16 & 6.35$\pm$0.44  & 4.20$\pm$0.21\\
                            & \emph{Completion Ratio} (\%) $\uparrow$ & 87.21$\pm$0.96 & 90.56$\pm$0.22 & 74.34$\pm$0.82 & 65.01$\pm$1.61 & 79.03$\pm$0.90\\
      \midrule
     \multirow{3}{*}{DSGT} & \emph{Artifacts} (cm) $\downarrow$ & 20.76$\pm$8.31 & 17.66$\pm$1.92 & N/A & N/A & N/A\\
                            & \emph{Holes} (cm) $\downarrow$ & 9.09$\pm$6.28 & 2.58$\pm$0.05 & N/A & N/A & N/A\\
                            & \emph{Completion Ratio} (\%) $\uparrow$ & 56.47$\pm$1.30 & 90.69$\pm$0.41 & N/A & N/A & N/A\\
      \midrule 
      \multirow{3}{*}{DiNNO} & \emph{Artifacts} (cm) $\downarrow$ & \color{DGreen}\textbf{2.78$\pm$0.34} & 5.78$\pm$1.50 & 7.56$\pm$2.62 & 61.29$\pm$7.41 & 19.35$\pm$2.97\\
                            & \emph{Holes} (cm) $\downarrow$ & 2.79$\pm$0.22 & 2.33$\pm$0.22 & 4.10$\pm$0.39 & 5.73$\pm$0.82& 3.74$\pm$0.41\\
                            & \emph{Completion Ratio} (\%) $\uparrow$ & 85.90$\pm$2.55 & 91.45$\pm$1.99  & 79.75$\pm$3.71 & 64.73$\pm$5.94 & 80.46$\pm$3.55\\
      \midrule 
    \multirow{3}{*}{RAMEN (ours)} & \emph{Artifacts} (cm) $\downarrow$ & 3.25$\pm$0.35 & \color{DGreen}\textbf{5.47$\pm$0.97} & \color{DGreen}\textbf{5.55$\pm$0.35}& \color{DGreen}\textbf{37.67$\pm$1.64}& \color{DGreen}\textbf{12.98$\pm$0.83}\\
                            & \emph{Holes} (cm) $\downarrow$ & \color{DGreen}\textbf{1.78$\pm$0.04} & \color{DGreen}\textbf{2.01$\pm$0.03} & \color{DGreen}\textbf{3.23$\pm$0.08}& \color{DGreen}\textbf{3.28$\pm$0.11}&\color{DGreen}\textbf{2.57$\pm$0.07}\\
                            & \emph{Completion Ratio} (\%) $\uparrow$ & \color{DGreen}\textbf{94.37$\pm$0.27} & \color{DGreen}\textbf{94.20$\pm$0.32} & \color{DGreen}\textbf{90.72$\pm$0.84}& \color{DGreen}\textbf{86.48$\pm$0.64}& \color{DGreen}\textbf{91.44$\pm$0.52}\\
      \bottomrule
    \end{tabular}%
}
  \label{tab:simtest_1}
\end{table*}

\begin{figure*}[hbt!]
	\centerline{\includegraphics[trim={0cm 8cm 0cm 0cm},clip,width=2\columnwidth]{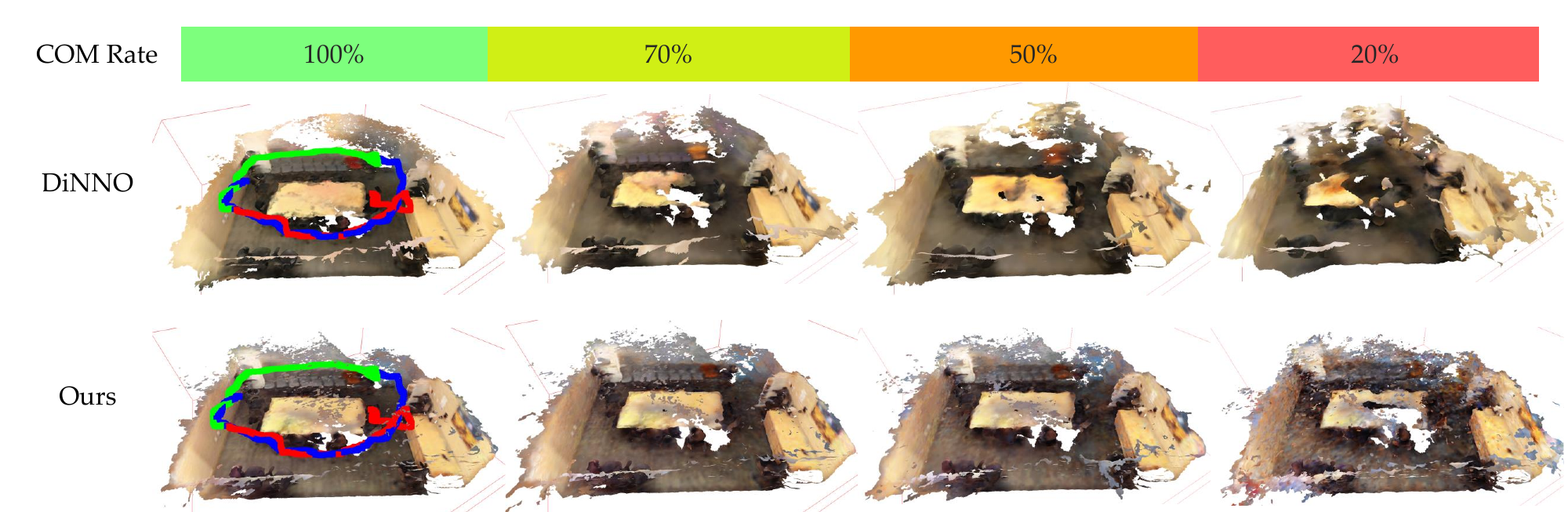}} 
	\caption{Reconstruction quality of DiNNO and RAMEN (ours) on ScanNet ``scene0169" under varying communication success rates (three-agent trajectories shown in different colors). Even at a 20\% success rate, RAMEN preserves scene details, while DiNNO exhibits incomplete or blurry reconstructions.}
	\label{fig:sim_test2}
\end{figure*}

\begin{figure*}[hbt!]
	\centerline{\includegraphics[trim={1.2cm 6.3cm 1.2cm 5cm},clip,width=2\columnwidth]{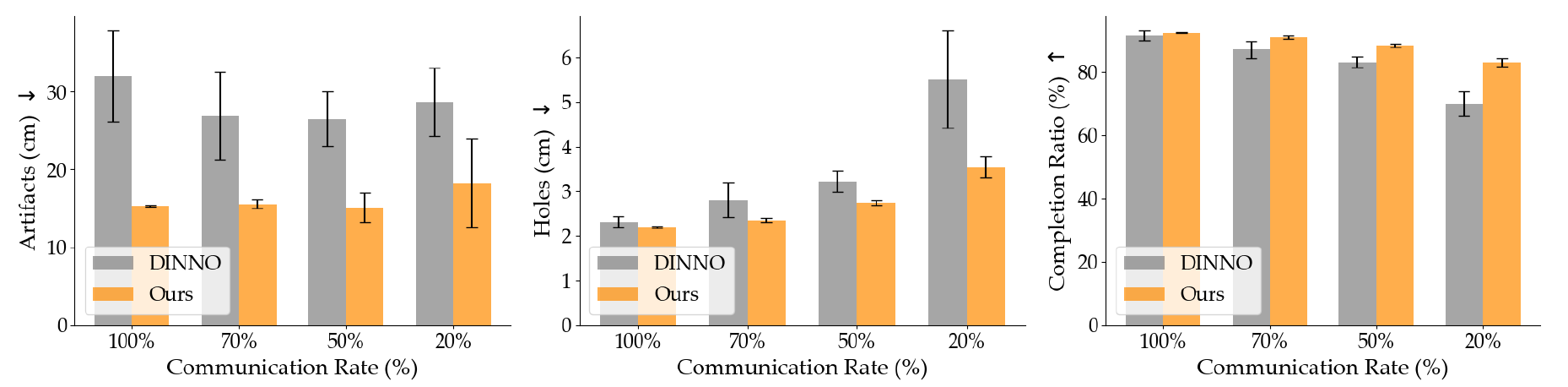}} 
	\caption{Evaluation metric scores for DiNNO and RAMEN (ours) on ScanNet ``scene0169" under varying communication success rates. While both methods exhibit worse scores with lower communication rate, RAMEN demonstrates greater robustness, maintaining higher accuracy and lower variance than DiNNO, especially in lower communication rates. }
	\label{fig:sim_test2_bars}
\end{figure*}

We reconstructed meshes for evaluation by querying the neural implicit maps within a uniform voxel grid and applying the marching cubes algorithm \cite{marching_cube}. 
To assess the accuracy of the reconstructed geometry, we employ the following 3D quantitative metrics, adapted and renamed for clarity from prior neural implicit mapping works \cite{iMAP,NICE-SLAM, eslam, xin2024heroslam}:
\begin{enumerate}
    \item{\emph{Artifacts} (cm):}
    The average distance between sampled points from the reconstructed mesh and the nearest ground-truth point. Higher values indicate more reconstruction artifacts. Thus, lower values represent better performance.
    \item{\emph{Holes} (cm):} 
    The average distance between sampled points from the ground-truth mesh and the nearest reconstructed point. Higher values indicate larger holes in the reconstruction. Thus, lower values represent better performance.
    \item{\emph{Completion Ratio} (\%):}
    The percentage of points in the reconstructed mesh with \emph{Holes} under 5 cm, with higher values representing better performance. 
    This metric provides a holistic measure of the captured scene geometry and is our primary evaluation criterion.
\end{enumerate}

\begin{figure*}[hbt!]
	\centerline{\includegraphics[trim={0cm 4cm 0.5cm 4cm},clip,width=2\columnwidth]{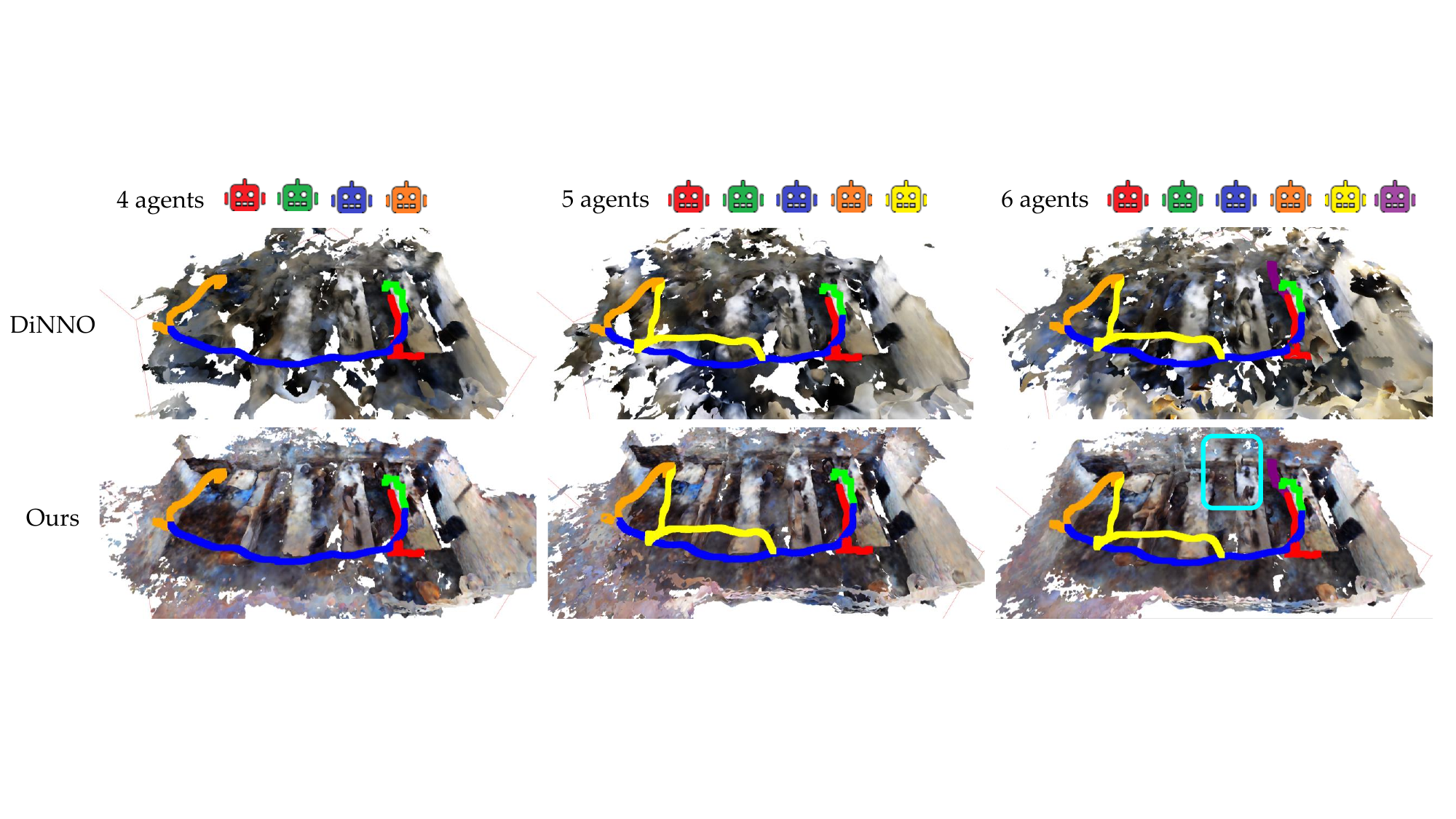}} 
	\caption{Reconstruction quality of DiNNO and RAMEN (ours) on ScanNet ``scene0106" with varying numbers of agents (trajectories shown in different colors). RAMEN effectively utilizes observations from additional agents to achieve a more complete scene reconstruction, particularly in the highlighted region (cyan box). In contrast, DiNNO exhibits numerous artifacts and incomplete reconstructions.}
	\label{fig:sim_test3}
\end{figure*}

\begin{figure*}[hbt!]
	\centerline{\includegraphics[trim={1.2cm 6.3cm 1.2cm 5cm},clip,width=2\columnwidth]{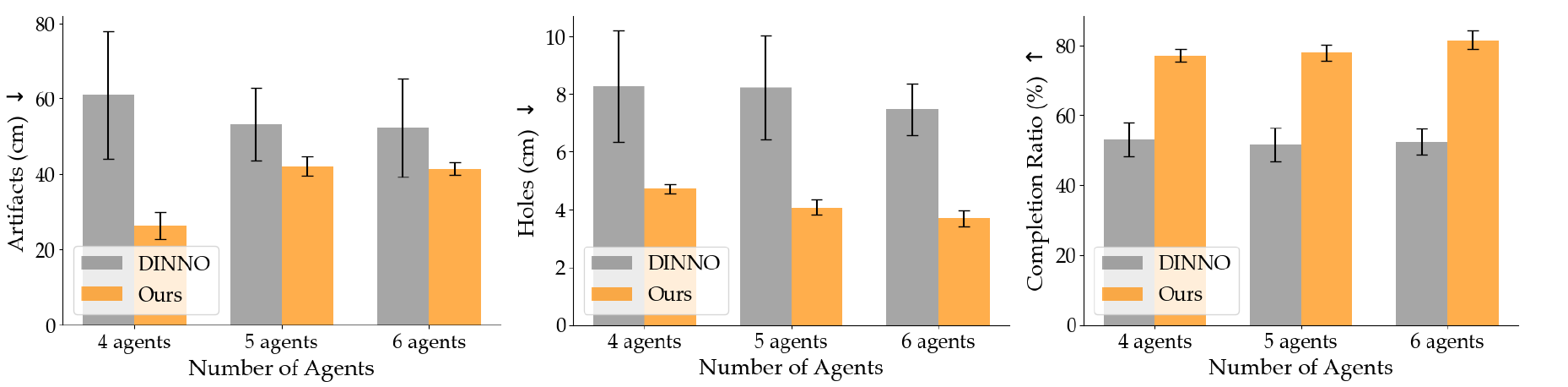}} 
	\caption{Evaluation metric scores for DiNNO and RAMEN (ours) on ScanNet ``scene0106" with varying numbers of agents. DiNNO exhibits worse artifact scores due to noisy reconstructions. While both methods benefit from additional agents, RAMEN more effectively leverages the increased information to achieve greater improvements in scene completion. }
	\label{fig:sim_test3_bars}
\end{figure*}

The metrics mentioned above can be combined to provide a more detailed picture of performance. For example, a high \emph{Completion Ratio} with high \emph{Holes} indicates small isolated objects missing from the reconstructed mesh.

\subsection{Simulated Experiments}

\textbf{\emph{How does RAMEN perform in general?}} 
We first investigate RAMEN's general performance with three agents, a 50\% communication success rate, and a fully-connected communication graph. 
We evaluated RAMEN and the baseline methods on four scenes: ``office1" and ``room1" from the Replica dataset, and ``scene0000" and ``scene0106" from the ScanNet dataset. 
For a comprehensive evaluation, we conducted three trials per agent per method on each scene and reported the mean and standard deviation of the performance metrics.
As shown in Table \ref{tab:simtest_1}, RAMEN consistently achieves the best overall performance. 
DiNNO only slightly outperforms RAMEN in one metric on the simple synthetic scene ``office1''.
While DSGD and DSGT perform adequately on the relatively simple Replica dataset, DSGT fails to converge on the more challenging ScanNet dataset under asynchronous communication. 
Notably, RAMEN surpasses DiNNO in completion ratio by a significant margin, demonstrating its ability to effectively combine reliable information from all agents to reconstruct a complete scene. 
Furthermore, RAMEN exhibits significantly smaller standard deviations compared to DiNNO, indicating a more stable and robust mapping process.  
Figure \ref{fig:sim_test1} provides a visualization example from ``scene0000," where RAMEN reconstructs the highlighted region with higher accuracy than DiNNO.
Since DiNNO achieves the best performance among the baseline methods, for the rest of the experiments, we only compared RAMEN to DiNNO.

We also emphasize that RAMEN employs a simple uncertainty quantifier, incurring minimal computation overhead compared to existing NeRF uncertainty methods covered in Section \ref{section:related works}. 
For instance, in a representative three-agent experiment conducted on an RTX 4090 GPU, RAMEN runs at 8.58 FPS, while the baseline DiNNO runs at 9.38 FPS.

\begin{figure*}[hbt!]
	\centerline{\includegraphics[trim={3cm 5cm 3cm 0cm},clip,width=1.8\columnwidth]{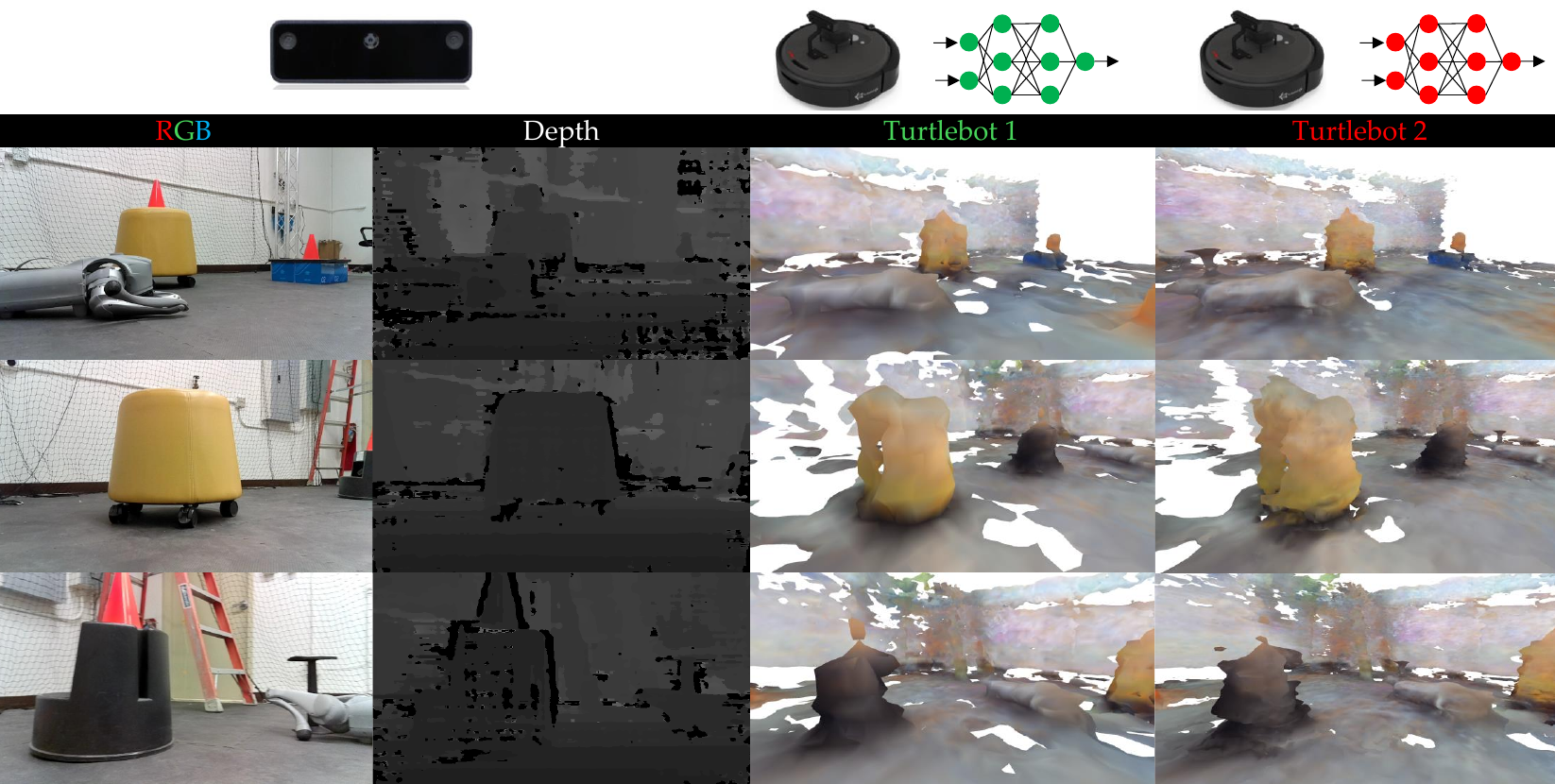}} 
	\caption{Comparison of RAMEN reconstructed scene details with corresponding captured images. 
    Each robot only physically visited half of the scene.
    The complete and similar reconstruction results from both robots indicate successful convergence to consensus. 
    Despite limitations in capturing fine details due to the low quality of depth images, both robots accurately learn the overall scene structure.}
	\label{fig:hardware_imgs}
\end{figure*}

\textbf{\emph{How does RAMEN perform with severe communication disruptions?}}
Next, we evaluate RAMEN under more challenging communication conditions. 
We conducted experiments with three agents exploring ``scene0169" from the ScanNet dataset, gradually decreasing the communication success rate from 100\% (synchronous updates) to 20\%.
Figure \ref{fig:sim_test2} shows that RAMEN preserves the overall structure and details of the scene even at a 20\% communication success rate, while DiNNO fails to reconstruct half of the scene. 
As expected, the performance metrics (shown in Figure \ref{fig:sim_test2_bars}) deteriorate for both methods as the communication success rate decreases. 
However, RAMEN exhibits significantly less performance degradation compared to DiNNO.  
Furthermore, consistent with the previous experiment, RAMEN demonstrates much smaller standard deviations, indicating greater robustness. 
Interestingly, RAMEN even outperforms DiNNO under synchronous communication.
These results highlight RAMEN's resilience under extreme communication conditions.

\textbf{\emph{How does RAMEN scale with more agents?}}
Finally, we investigate the scalability of RAMEN. 
We evaluated RAMEN and DiNNO on ``scene0106" from the ScanNet dataset with 4, 5, and 6 robotic agents under a 50\% communication success rate. 
To assess information propagation through incomplete communication graph, we restricted agents to communicate only with their immediate neighbors.
For instance, with 4 agents, the possible communication among agent $0,1,2,3$ are $\mathcal{E} = \{ (0,1), (1,2), (2,3) \}$.
Figure \ref{fig:sim_test3} demonstrates RAMEN's effectiveness in utilizing observations from additional agents. 
For example, in the highlighted region (cyan box), RAMEN successfully reconstructs the computer on the desk with the extra input from the $6^{\text{th}}$ agent. 
In contrast, DiNNO struggles to represent the complete scene and produces numerous floating artifacts, leading to high artifact scores, as shown in Figure \ref{fig:sim_test3_bars}. 
Figure \ref{fig:sim_test3_bars} also demonstrates that RAMEN achieves better scene completion with additional agents, further highlighting the effectiveness of our uncertainty-based weights in combining information from neighboring agents.

\subsection{Real-World Hardware Experiments}
To evaluate RAMEN's performance in real-time multi-robot mapping, we used two Turtlebots within the Robot Operating System (ROS)~\cite{ROS2}. Each robot was manually controlled to explore half of the environment, capturing RGBD images with its onboard sensor. 
Figure \ref{fig:teaser} illustrates the trajectories and areas covered by each robot. In this setup, due to the limited range of the stereo camera, successful consensus between the robots is crucial for reconstructing the complete scene.
We used a VICON motion capture system to provide ground-truth robot poses. 
Images and pose data from each Turtlebot were streamed to a separate PC via ROS2 and WiFi for real-time processing, with mapping optimization performed at around 1Hz. 
Two PCs exhibited substantial differences in processing speed, and \emph{no} synchronization mechanism was implemented to align their optimization iteration counts.
Due to bandwidth limitations, the two PCs could only exchange their neural implicit maps every 30 seconds, resulting in a communication success rate of 3.33\%. 
This asynchronous communication setup, combined with noisy depth images and motion blur, presents an \textit{extremely} challenging scenario that thoroughly tests RAMEN's robustness.

\begin{table}[!hbt]
    \caption{Hardware experiment results}
    \centering
    \resizebox{\columnwidth}{!}{%
        \begin{tabular}{ccccc}
            \toprule
            Robot& Method & \emph{Artifacts} (cm) $\downarrow$ & \emph{Holes} (cm) $\downarrow$ & \makecell{\emph{Completion}\\\emph{Ratio} (\%) $\uparrow$} \\
            \midrule
            \multirow{2}{*}{Turtlebot 1} & DiNNO & 26.43 & 4.70 & 63.07\\
                                         & RAMEN & \color{DGreen}\textbf{6.50} & \color{DGreen}\textbf{4.43} & \color{DGreen}\textbf{78.47}\\
            \midrule
            \multirow{2}{*}{Turtlebot 2} & DiNNO & 10.76 & 12.77 & 39.83\\
                                         & RAMEN & \color{DGreen}\textbf{5.87} & \color{DGreen}\textbf{5.15} & \color{DGreen}\textbf{69.66}\\
            \midrule
        \end{tabular}%
}
    \label{tab:hardware}
\end{table}

Figure \ref{fig:teaser} shows the overall mapping results.
For comparison, we generated a ground-truth reconstruction using a single robot that captured 4x more images while exploring the entire scene. 
Under these challenging conditions, DiNNO fails to converge, while RAMEN enables both Turtlebots to reach consensus and reconstruct the complete scene. 
Figure \ref{fig:hardware_imgs} provides a closer look at the reconstructed details. 
Despite noisy depth images, the overall geometry is accurately captured. 
The absence of floating artifacts, in contrast to the simulated experiments, is attributed to human-in-the-loop active perception, which helped to minimize such errors. 
This also explains the higher metric scores in Table \ref{tab:hardware} compared to the previous, less challenging simulated experiments.

\section{Limitations}
A key limitation of the proposed frequency-based epistemic uncertainty is its inability to differentiate between regions with varying geometric complexity. 
Regions with complex geometries, such as those containing many small objects or fine details, require more observations to reduce uncertainty and improve reconstruction quality.  
Another limitation, inherited from Co-SLAM, is the reliance on a keyframe database $R_i$.
This necessitates storing a large number of images onboard and continuously replaying them to prevent catastrophic forgetting, posing challenges for large-scale mapping tasks.
Our reconstruction quality also suffer from significant depth estimation noise introduced by the RGBD camera on the Turtlebot. 
Finally, controlling the robots manually in our hardware experiments may have introduced a human bias, potentially affecting reproducibility.
A promising future direction is to integrate active neural implicit mapping techniques, such as those proposed in \cite{ActiveNeuralMap}, to automate data acquisition.

\section{Conclusion} 
This paper introduces RAMEN, a real-time asynchronous multi-agent neural implicit mapping framework.  
By utilizing a weighted version of C-ADMM, RAMEN handles communication disruptions and enables robots to reconstruct a complete scene collaboratively.  
The consensus optimization in RAMEN prioritizes information from neighboring robots with lower uncertainty, leading to more accurate and robust mapping.  
Extensive simulation and hardware experiments demonstrate the effectiveness of RAMEN in challenging scenarios.

\section*{Acknowledgments}
This work is supported in part by the National Science Foundation, under grants ECCS-2438314 CAREER Award, CNS-2423130, and CCF-2423131.


\bibliographystyle{ieeetr}
\bibliography{references}

\end{document}